\documentclass[12pt]{iopart}

\usepackage{iopams} 
\usepackage{url} 
\usepackage{lineno}
\usepackage[inline]{trackchanges} 
\usepackage{soul}
\usepackage{amsmath}
\usepackage{graphicx}
\usepackage{booktabs}
\usepackage{amssymb}
\usepackage{multirow}
\usepackage{dsfont}
\usepackage{hyperref}
\usepackage{xcolor}
\usepackage{color}
\usepackage[style=authoryear, maxcitenames=2, natbib=true]{biblatex}
\newcommand{\ie}{\textit{i.e., }}

\newcommand{\networkA}{LSTM}
\newcommand{\networkB}{\textit{Index-LSTM}}
\newcommand{\networkC}{\textit{ViT-LSTM}}
\newcommand{\stows}{\texttt{S2S}}

\begin{document}
\title[Deep Learning Meets Teleconnections: Improving S2S Predictions]{Deep Learning Meets Teleconnections: Improving S2S Predictions for European Winter Weather}
%
%




\author{Philine L. Bommer$^{1,2}$, Marlene Kretschmer$^{3,4}$, Fiona R. Spuler$^{4,5}$, Kirill Bykov$^{1,2,7}$, Marina M.-C. H\"ohne$^{1,2,6,7}$}

\address{$^1$Understandable Machine Intelligence Lab, TU Berlin, Berlin, Germany}
\address{$^2$Department of Data Science, ATB, Potsdam, Germany}
\address{$^3$Leipzig Institute for Meteorology, Leipzig University, Leipzig, Germany}
\address{$^4$Department of Meteorology, University of Reading, Reading, UK} 
\address{$^5$The Alan Turing Institute, London, UK}
\address{$^6$Institute of Computer Science - University of Potsdam, Potsdam, Germany}
\address{$^7$BIFOLD – Berlin Institute for the Foundations of Learning and Data, Berlin,
Germany}





\eads{\mailto{pbommer@atb-potsdam.de}}



%
%

%
%

\begin{abstract}
Predictions on subseasonal-to-seasonal (\stows) timescales—ranging from two weeks to two months—are crucial for early warning systems but remain challenging owing to chaos in the climate system. Teleconnections, such as the stratospheric polar vortex (SPV) and Madden-Julian Oscillation (MJO), offer windows of enhanced predictability, however, their complex interactions remain underutilized in operational forecasting.
Here, we developed and evaluated deep learning architectures to predict North Atlantic-European (NAE) weather regimes, systematically assessing the role of remote drivers in improving \stows\ forecast skill of deep learning models. We implemented (1) a Long Short-term Memory (LSTM) network predicting the NAE regimes of the next six weeks based on previous regimes, (2) an Index-LSTM incorporating SPV and MJO indices, and (3) a ViT-LSTM using a Vision Transformer to directly encode stratospheric wind and tropical outgoing longwave radiation fields. These models are compared with operational hindcasts as well as other AI models. Our results show that leveraging teleconnection information enhances skill at longer lead times. Notably, the \networkC\ outperforms ECMWF's subseasonal hindcasts beyond week 4 by improving Scandinavian Blocking (SB) and Atlantic Ridge (AR) predictions. Analysis of high-confidence predictions reveals that NAO$-$, SB, and AR opportunity forecasts can be associated with SPV variability and MJO phase patterns aligning with established pathways, also indicating new patterns.
Overall, our work demonstrates that encoding physically meaningful climate fields can enhance \stows\  prediction skill, advancing AI-driven subseasonal forecast. Moreover, the experiments highlight the potential of deep learning methods as investigative tools, providing new insights into atmospheric dynamics and predictability.
\end{abstract}

%
\vspace{2pc}
\submitto{Machine Learning: Earth}
%
%

\section{Introduction}
Extreme weather events are becoming increasingly frequent and severe, posing a significant threat to societies, economies, and ecosystems. Improving longer lead-times predictions is essential for early warning systems and disaster mitigation, helping to reduce economic damage and humanitarian losses \citep{coughlan_de_perez_action-based_2016}. However, such subseasonal-to-seasonal (\stows) forecasts—spanning two weeks to two months— remain strongly limited in skill due to the chaotic nature of the climate system \citep{Vitart2017,vitart2018sub}.
Predictability at \stows timescales arises due to teleconnections, where anomalies in one region can influence the persistence and transition of weather patterns in distant locations \citep{ vautard1990multiple, Seager2010, nielsen_forecasting_2022} through wave propagation and large-scale circulation shifts \citep{yamagami_subseasonal_2020}.
In the North Atlantic–European (NAE) sector, the influence of teleconnections can be analyzed through the occurrence of weather regimes such as the positive (NAO+) and negative phase (NAO-) of the North Atlantic Oscillation, Atlantic Ridge (AR), and Scandinavian Blocking (SB). These persistent circulation patterns shape regional weather including extreme events \citep{vautard1990multiple,Cattiaux2010,Seager2010,luo2020combined,ardilouze2021flow, nielsen_forecasting_2022}. 
Among the key drivers of these regimes is the stratospheric polar vortex (SPV), a persistent cyclonic circulation in the Arctic stratosphere. 
Sudden stratospheric warming---rapid temperature increases in the polar stratosphere coinciding with a weakened flow---can disrupt the SPV, shifting the jet stream southward and triggering prolonged cold spells over northern Eurasia \citep{kretschmer2018more, domeisen_role_2020,spaeth2024stratospheric}.
Similarly, the Madden-Julian Oscillation (MJO), an eastward-propagating tropical convective system, has been shown to modulate the NAE regimes, e.g. influencing the likelihood of negative (NAO$-$) and positive North Atlantic Oscillation (NAO$+$) phase up to $20$ days
later \citep{cassou_intraseasonal_2008, Lee2019, lee2020links, robertson_subseasonal_2020, nardi_skillful_2020}. 
The variability in these teleconnection drivers, thus, offers windows of enhanced predictability that can be leveraged to improve extreme weather forecasts \citep{mariotti_windows_2020}.

Despite recent research progress, our understanding of teleconnections remains limited, largely due to the complexity of their interactions \citep{kretschmer2017early}. In particular, it has been shown that most numerical weather models might underestimate established teleconnection pathways, e.g. struggling with ENSO variability as well as MJO phases influencing European weather or capturing stratosphere-troposphere couplings and the SPV \citep{robertson_subseasonal_2020, Spaeth2024, riviere2024opposite,wu2024tropospheric, garfinkel2025process}.
Consequently, operational forecast systems struggle to accurately capture climate and weather dynamics beyond two weeks, limiting their utility for impact-driven decision-making and extreme weather preparedness. 

Machine learning approaches, particularly deep learning (DL), are showing promise not only in the weather domain but also in the \stows\  domain. While early efforts focused on convolutional recurrent models for spatiotemporal precipitation forecasting \citep{shi_convolutional_2015}, later studies have already discussed the general potential of deep learning in \stows\  prediction \citep{cohen2019s2s}. Progress has continued with the introduction of generative neural weather models \citep{weyn_sub-seasonal_2021} and explainable AI (XAI) frameworks even identifying potential teleconnections and windows of enhanced predictability \citep{mayer_subseasonal_2020}. In recent years, works like \citet{castro_stconvs2s_2021},  \citet{peng_polar_2021}, and \citet{mouatadid_adaptive_2022} have demonstrated the potential of hybrid data-driven models for teleconnection-aware prediction, physics-informed networks, and ML-driven bias correction of \stows\  forecasts. Advances include transformer-based architectures for extended-range forecasts \citep{zhang2024adapting} and multi-modal approaches combining physical and data-driven information \citep{perez2024earth}, underscoring a growing trend toward more interpretable and skillful DL tools in \stows\  forecasting. Nonetheless, ML for \stows\  forecasting remains in its infancy, facing two fundamental challenges: first, capturing multi-timescale climate dynamics, and second, effectively representing teleconnections within ML architectures.
To address these challenges and systematically evaluate the role of teleconnection drivers in \stows\ prediction skill, we develop DL approaches to forecast NAE regimes in boreal winter. We develop and compare three architectures with increasing complexity: 
\begin{enumerate}
    \item a basic Long Short-term Memory network (\networkA)---an LSTM trained to predict NAE regimes for six weeks, using only the past six weeks of regime states as input.
    \item An index-augmented LSTM (\networkB)---an extension of \networkA\ by incorporating physical driver indices—the stratospheric polar vortex (SPV) strength and the Madden-Julian Oscillation (MJO) phase.
    \item A spatiotemporal model (\networkC)---an encoder-decoder model, consisting of a Vision Transformer (ViT) and an LSTM model that integrates the raw climate fields of zonal winds at 10 hPa (u10) over the polar region and outgoing longwave radiation (OLR) over the tropics—to allow the network to directly learn spatiotemporal teleconnection information.
\end{enumerate}

Using the well-established NAE regime framework, where MJO and SPV variability play a key role in predictability \citep{spuler2025}, we demonstrate that physically meaningful drivers can enhance the \stows\ forecast skill. Moreover, we compare our predictions with other established machine learning approaches (Aurora-based architecture \citep{bodnar2024aurora} and logistic regression), as well as the dynamical hindcasts from the ECMWF subseasonal forecast model, which relies on the numerical implementation of physical laws.
Finally, we highlight the investigative potential of machine learning by analyzing the precursor patterns used by the network to make high-probability predictions. By bridging data-driven forecasting with a process-based climate understanding, our study contributes to advancing \stows\  predictability.

Our work is structured as follows: In Section \ref{sec:data} we provide a description of the data used and the applied preprocessing steps. The different network architectures, baselines, and evaluation measures are discussed in Section \ref{sec:methods}. In Section \ref{sec:results} we detail the forecast skill results (Section \ref{sec:fore-skill}) and the analysis of the precursor and teleconnection patterns (Section \ref{sec:woo}). Finally, in Section \ref{sec:discuss}, we discuss our results and limitations and our conclusion.

\section{Data \& Processing}
\subsection{Data}\label{sec:data}

We use daily-mean ERA5 data \citep{Hersbach2020} in the satellite period from $1980$ to $2023$, providing $43$ years of reanalysis data which we treat as observations. For the Vision transformer training, which requires a larger training set and focuses on the spatial patterns of the climate variables, we complement this with daily-mean 20CRv3 reanalysis data \citep{Slivinski2019} from $1836$ to $1980$, adding $144$ years of reanalysis data. 

Our study focuses on the extended boreal winter (16 November to 31 March) of the following climate variables: geopotential height data at $500$ hPa (z500) over the North Atlantic region ($90^{\circ}$W–$30^{\circ}$E, $20^{\circ}$–$80^{\circ}$N), zonal winds at $10\text{ hPa}$ (u10) over the polar region ($60^\circ - 90^\circ$N), and outgoing longwave radiation (olr) in the tropics ($15^\circ$S - $15^\circ$N) (see Table \ref{tab:variables}). 

All variables are first regridded, resulting in a $22 \times 256$ grid for u10 and olr (see Appendix \ref{app:data}). To reduce short-term variability, we apply a rolling seven-day mean resulting in $137$ weekly mean data points per winter, i.e., in total $2236$ weekly samples for ERA5 and $7294$ 20CRv3 samples. 
Anomaly maps are then computed by subtracting the daily climatology from the daily data, with the climatology computed as the mean over the previous 30 years of the corresponding day of the year. For example, the anomaly of January 1st, $2010$ is computed by subtracting the mean of all 1st January days $1980-2009$ (see \citep{world2017wmo} for the specific procedure). For days in the period $1980-2009$ (where a preceding $30$-years period is not in our dataset), the corresponding daily climatology is computed using data from $1980$ to $2009$. 

To compare the performance of our ML-based models with operational systems, we use hindcasts from the CY47R3\_LR experiment \citep{Roberts2023}, which implements the 47r3 cycle of the ECMWF IFS in a lower-resolution setting over 1980-2020. 
Despite its lower resolution, the forecast skill for the studied regimes was shown to be comparable to the operational higher-resolution setting \citep{Roberts2023}. 
The dataset consists of 11-member ensemble forecasts of geopotential height at 500hPa initialized on the 1st, 8th,
15th and 22nd of each month with lead times of up to 47 days. Data was downloaded at a resolution of $2.5^\circ\times2.5^\circ$ for all dates covering the extended winter months studied. The hindcasts are preprocessed analogous to ERA5 data, applying a seven-day rolling mean first, and then computing anomalies separately for each lead time.

\subsection{North Atlantic European (NAE) regimes}\label{sec:regimes}
To compute the NAE regimes in the reanalysis data, we follow the methodologies presented in \citet{michelangeli1995weather}, \citet{cassou_intraseasonal_2008}, \citet{hannachi2017} and \citet{nielsen_forecasting_2022}. First, we apply a dimensionality reduction step by computing the empirical orthogonal function (EOF) components (which are equivalent to principal component analysis - PCA) of the z500 anomalies over the North Atlantic European region. We use the first $14$ EOFs, capturing $\geq94$\% of the total variance \citep{bloomfield2018changing, van2019influence}. 

\begin{figure}[t!]
\includegraphics[width=\textwidth]{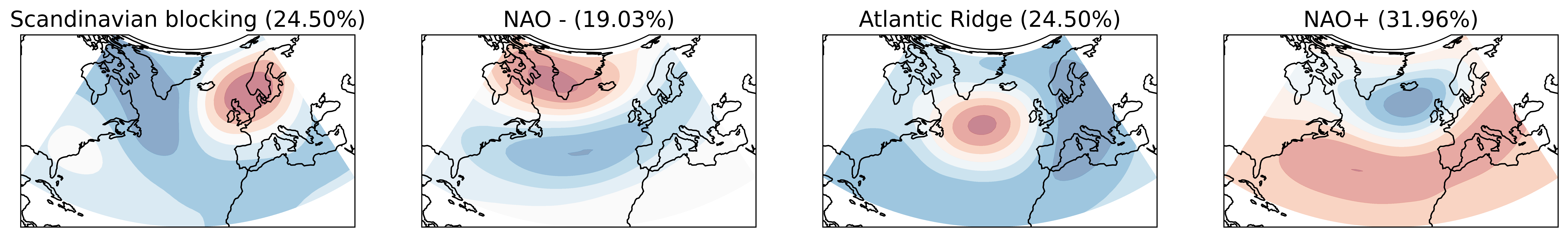}
    \caption[]{Maps of the NAE regimes computed for ERA$5$ data between $1980$ to $2023$ (see \citet{hannachi2017} for comparison). The regime name and percentage of occurrence within the data are provided in the panel titles.}
    \label{fig:nae-regimes}
\end{figure}
The EOFs of the data are then clustered into four groups using the k-means-clustering algorithm \citep{lloyd1982least}, resulting in the four NAE regimes. These regimes, characterized by their mean composites correspond to well-established patterns (see Figure \ref{fig:nae-regimes}): the positive phase of the North Atlantic Oscillation ($\text{NAO}+$, $30\%$ of all weeks), the negative phase of the North Atlantic Oscillation ($\text{NAO}-$, $19\%$ of all weeks), Scandinavian Blocking (SB, $24\% $ of all weeks), and Atlantic Ridge (AR, $26\% $ of all weeks). The identified regime frequencies and spatial patterns closely align with those reported in \citet{cassou_intraseasonal_2008}, with expected variations due to differences in the analysis period. We extend the regime analysis to hindcasts, by projecting the z500 data onto the ERA5 EOFs. The first 14 EOFs are then assigned to the ERA5 cluster centroids to compute the corresponding regimes (see also Appendix \ref{app:data}).

\subsection{Teleconnection drivers}\label{sec:data-indicies}
In addition to the spatial fields of u10 and olr anomalies, we compute indices capturing the strength of the stratospheric polar vortex (SPV) and the phase of the Madden–Julian Oscillation (MJO). The weekly SPV index is derived from the preprocessed u10 data by averaging over $60^{\circ}$N \citep{domeisen_role_2020}. For the MJO, we use the daily MJO index times series from $1979$ to $2023$ provided by NOAA, consisting of the first two principal components (RMM1 and RMM2) of combined tropical variables \citep{kiladis2014comparison}. Specifically, RMM 1 and RMM2 are calculated as the first two EOFs of the olr, 850-hPa zonal winds, and 200-hPa zonal wind across the tropics. To calculate the weekly MJO phases, we apply a seven-day rolling mean to both components and then compute the corresponding amplitude and phase index across eight MJO phases following \citet{wheeler2004all}. Active MJO phases (with amplitude $\geq1$) are assigned to classes $1-8$, whereas inactive phases (amplitude $<1$) are grouped into a separate class $0$ to account for the difference in teleconnection strength \citep{Lee2019}.

\section{Methods}\label{sec:methods}

We develop three DL-based architectures for \stows-scale prediction of NAE regimes, varying in complexity and input types.
As schematically shown in Figure \ref{fig:architectures}, each network is trained to forecast the NAE regimes for the next 6 weeks $T = [t+1,t+6]$ given the NAE regimes from the preceding six weeks $T = [t-5,t]$. The input $\mathbf{x}\in \mathbb{R}^{6\times 4}$, thus, consists of NAE regime classes over the past six weeks, where each class is represented as a one-hot encoded vector of length four. The output of the model $\mathbf{y} \in \mathbb{R}^{6\times 4}$ contains a probability distribution over the four NAE regimes for each predicted week. 

\subsection{LSTM}\label{sec:lstm-network}
We first use a Long-short-term memory (LSTM) network (see Figure \ref{fig:architectures}A), which is well-suited for capturing temporal dependencies in limited datasets \citep{hochreiter1997long}. The input consists solely of the NAE regime classes over the previous six weeks, which are processed by coupled LSTM cells in a sequence-to-sequence architecture (see Appendix \ref{app:architectures} for details). The output layer is a time-distributed linear layer that predicts weekly class probabilities.

\subsection{Index-LSTM}\label{sec:index-network}
We extend the LSTM model by integrating remote teleconnection drivers - the SPV index \citep{domeisen_role_2020} and MJO phase index \citep{wheeler2004all}.
As shown in Figure \ref{fig:architectures} B, the \networkB\ receives an augmented input vector $\hat{\mathbf{x}}\in\mathbb{R}^{6\times13}$, which concatenates the one-hot encoded NAE regimes ($\mathbb{R}^{6x4}$), real-valued SPV ($\mathbb{R}^{6x1}$) and one-hot encoded vector MJO phase index ($\mathbb{R}^{6x9}$).

\subsection{ViT-LSTM}\label{sec:s-t-network}
\begin{figure}
\includegraphics[width=\textwidth]{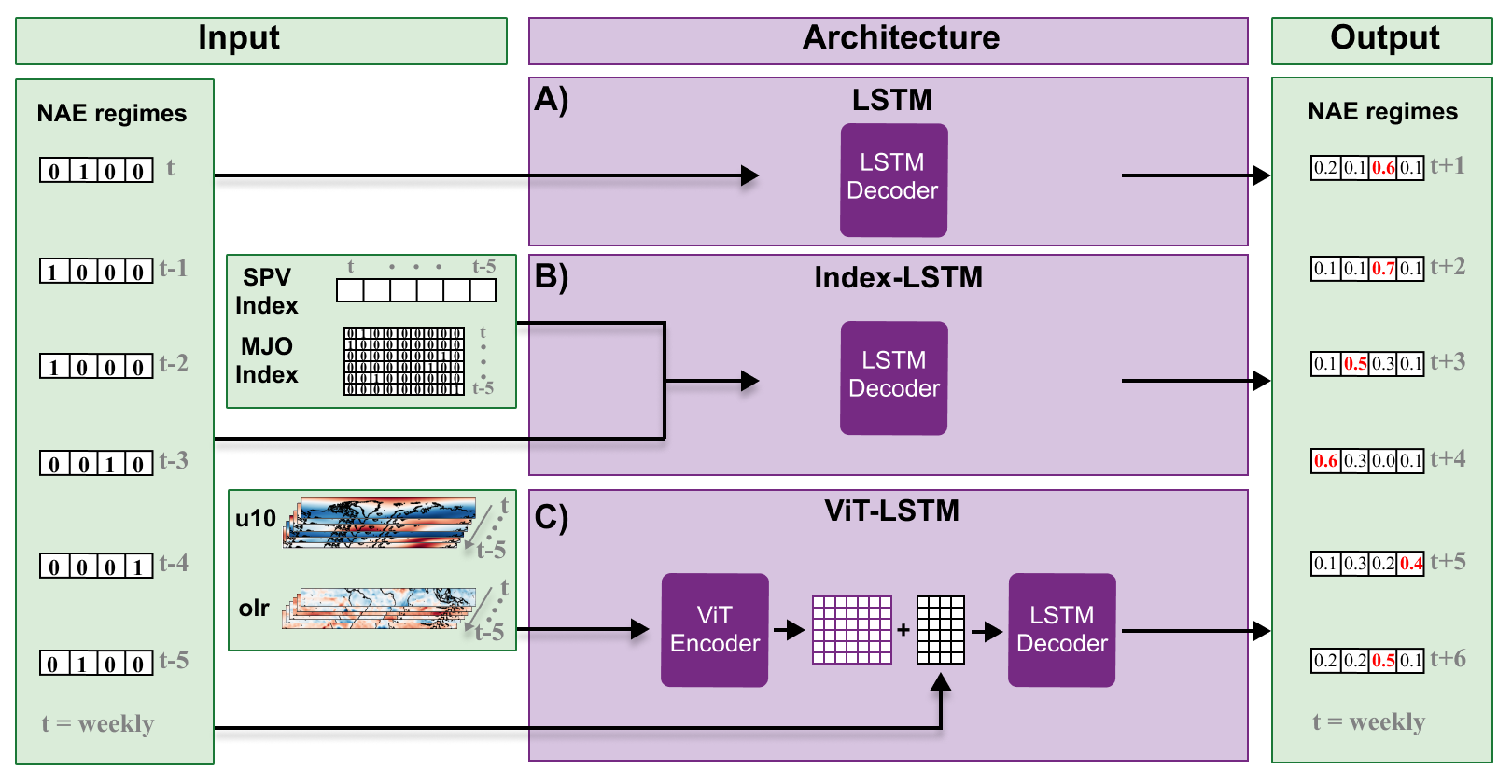}
    \caption[Architectures]{Schematic of the three DL architectures, arranged vertically by increasing complexity: A) \networkA, B) \networkB, and C) \networkC. Each model is designed to predict the probabilities of North Atlantic-European (NAE) regimes for the next six weeks, given NAE regime sequences from the previous six weeks (input: left panel; output: right panel). \networkA\ only uses NAE regime classifications as input. 
    \networkB\ extends this approach by incorporating additional dynamical predictors: the Madden-Julian Oscillation (MJO) phase index as a one-hot encoded vector and the stratospheric polar vortex (SPV) index as a continuous variable (additional input panel in B). \networkC\ represents the most advanced architecture, replacing predefined teleconnection indices with spatial climate fields. Instead of receiving MJO and SPV indices, it directly processes zonal wind (u10) in the polar region and the outgoing longwave radiation (olr) in the tropics, both for the past six weeks. These fields are encoded using a Vision Transformer (ViT) encoder, which extracts spatial features. These are then combined with the regime class information and passed to the LSTM-decoder, enabling the model to learn from spatial information potentially not captured by the conventional indices that influence \stows\ regime variability.}
    \label{fig:architectures}
\end{figure}

Since \networkB\ only relies on predefined, and rather simple indices to capture known SPV and MJO teleconnections, the model potentially misses the relevant spatio-temporal information. Thus, we construct a third model, that combines the spatial anomalies fields of u10 and olr with the NAE regimes. By integrating these spatial fields instead of the indices, ViT-LSTM enables the model to autonomously learn the relevant driver patterns, assuming that they are not optimally captured by conventional indices. 
More precisely, as illustrated in Figure \ref{fig:architectures}C we extend the \networkA\ by adding two combined Vision Transformers (ViT) as the encoder \citep{dosovitskiy2020image}. We train each ViT using a masked autoencoder (MAE) setup \citep{He2022}. By reconstructing masked patches in weekly u10 or weekly olr anomaly maps (see also Figure \ref{fig:A-vit-lstm}), each ViT learns to decode relevant climate patterns (for further details on MAE pre-training, and hyperparameters see \ref{app:architectures}). We extract only the ViT encoder of each MAE setup and combine them. Thus, in the pre-trained encoder, each ViT encodes six weeks of u10 and olr fields into an embedding vector, extracting spatial patterns. The extracted embeddings (violet array in Figure \ref{fig:architectures}C) are concatenated with NAE regime class information (black array) before being passed to the LSTM decoder.
To avoid overfitting due to limited training data, we apply dropout \citep{srivastava2014dropout} and batch normalization \citep{ioffe2015batch} to the embeddings. The decoder follows the same structure as \networkA, predicting class probabilities over the next six weeks.

\subsection{Training}\label{sec:training}
To train and evaluate each model, we split the ERA5 dataset into training (November $1980$ to March $2006$), validation (November $2006$ to March $2012$), and test (November $2012$ to March $2023$) sets. The chronological split prevents information leakage by avoiding overlapping winter weeks, thus ensuring that the validation and test sets contain distinct climate patterns while maximizing the number of data points (see \ref{app:training} for more details). We also tested an alternative random dataset assignment of boreal winters ($80\%$ training and $20\%$ test) and observed no significant impact on the performance. 

For the \networkC, which includes a ViT encoder, we employ a two-stage approach. First, we pre-train each ViT using a masked autoencoder (MAE) setup to reconstruct masked patches of weekly olr or u10 images. Given the complexity of the model (i.e., $\geq 5 \times 10^6$ parameters), for this step, we use both the ERA5 training set and 20CRv3 data (November 1836 to March 1969). We test the pre-trained ViTs on 20CRv3 data from November 1970 to March 1980 to assess their performance (i.e., reconstruction error). Secondly, after pre-training, we fine-tune the model, that is, the ViT encoders are frozen (non-trainable parameters), and only the decoder, dropout, and batch normalization layers are trained on the classification task. 
While \networkC\ requires the pre-training step, both \networkA\ and \networkB\ are trained directly on the classification task. 
Deeper networks have shown to be prone to miscalibration, limiting the usability of the classification probabilities at the output \citep{mukhoti2020calibrating}. Since we aim to predict the probabilities for each regime, we calibrate all three architectures during training. In addition, the hyperparameters are optimized using Bayesian optimization (BO) \citep{snoek2012practical} (see \ref{app:training} and Table~\ref{tab:hps} for details).

\subsection{Baselines \& additional Models}\label{sec:baselines}
To benchmark our architectures, we compare their performance against different forecasts.
\begin{itemize}
\item Persistence---Assumes that the last observed regime remains unchanged, thus the regime in week $t+1$ is the same as the regime in week $t$).
\item Climatology---For each week in the training set, the dominant NAE regimes (i.e., the most frequently occurring regime on the corresponding day over 30 years) are used as the deterministic climatological forecast.
\item Hindcasts---We computed the regimes by regridding the ERA5 z500 data to match the hindcast resolution (\ie $2.5^{\circ}$) and project both ERA5 and hindcast z500 anomalies onto ERA5-derived EOFs (see Appendix \ref{app:evaluation}).
\end{itemize}
Beyond these baselines, we evaluate additional machine learning and deep learning models:
\begin{itemize}
    \item Logistic regression (LR)---a widely used method across different forecast tasks \citep{gagne2017storm, jergensen2020classifying}, which predicts only based on the NAE regime time series.
    \item Aurora-T---an encoder-decoder neural network leveraging a pre-trained foundation model (Aurora, \citet{bodnar2024aurora}) as an encoder and a transformer decoder \citep{waswani2017attention}. 
    Unlike LR, the Aurora-T model also integrates spatial climate data (u10 and olr), enabling a more direct comparison with our proposed \networkC\ model.
\end{itemize}

\subsection{Skill Evaluation}
Since classifying the NAE regimes is an imbalanced multi-class problem, the standard accuracy can be misleading, as it tends to be dominated by the most frequent classes. To mitigate this issue, we use balanced accuracy \citep{kelleher2020fundamentals}, which computes the per-class recall and then averages it across all classes:
\begin{align}
    &\mathrm{Accuracy} = \frac{1}{S}\sum^C_{c=1}\frac{\mathrm{TP}(c)}{\mathrm{TP}(c)+\mathrm{FN}(c)},
    \label{eq:acc}
\end{align}
where $\text{TP}(c)$ and $\text{FN}(c)$ refer to the number of true positives and false negatives for class $c$, and $S$ is the total number of samples. 
To further quantify the predictive performance, we compute the critical success score (CSI) \citep{schaefer1990critical}, also known as the threat score, which measures the proportion of correctly predicted regimes relative to the total number of relevant instances to indicate the number of successful warnings: 
\begin{align}
    &\mathrm{CSI} = \frac{\mathrm{TP}}{\mathrm{TP}+\mathrm{FP}+\mathrm{FN}},
    \label{eq:csi}
\end{align}
where $\text{FP}$ represents the number of false positives.  Since CSI is originally defined for binary classification we apply a one-vs-all approach for each class and compute the weighted average across all classes\citep{nielsen_forecasting_2022}. To gain a deeper understanding of the regime-specific model behavior, we additionally evaluate the class-wise CSI and class-wise accuracy, where each class is evaluated in a one-vs-all classification setting.

\section{Experiments \& Results}\label{sec:results}
In the following, we evaluate the forecast skill of our models to assess the impact of integrating external driver information (SPV and MJO patterns) and to identify forecasts of opportunities where these drivers enhance predictability.
To ensure statistical robustness, we employ a deep ensemble approach \citep{lakshminarayanan2017simple} by training $100$ instances of each DL-based model with different random seeds. The mean and standard deviation across these $100$ ensemble members provide an estimate of the forecast uncertainty.

\begin{table}[t!]
\begin{center}
    \resizebox{\textwidth}{!}{ 
\begin{tabular}{llllllll}
\toprule
{}&{} & {lead week 1} & {lead week 2} & {lead week 3} & {lead week 4} & {lead week 5} & {lead week 6}\\
\midrule
\multirow{3}{*}{\textbf{Baseline}}&Persistence & $54.1\%$ & $31.7\%$ & $25.3\%$ & $22.3\%$ & $16.0\%$ & $23.6\%$  \\
&Climatology & $24.8\%$ & $24.6\%$ & $24.3\%$ & $23.6\%$ & $23.6\%$ & $22.6\%$ \\
&Hindcast & $\mathbf{66\%}$ & $\mathbf{46.1\%}$ & $\mathbf{36.5\%}$ & $\mathbf{33.4\%}$ & $29.5\%$ & $28.9\%$ \\
\midrule
\multirow{5}{*}{\textbf{ML}}& LR & $35.65\pm0.09\%$ & $32.53\pm0.08\%$ & $26.89\pm0.09\%$ & $20.46\pm0.09\%$ & $16.1\pm0.1\%$ & $23.30\pm0.08\%$\\
& LSTM & $39\pm3\%$ & $28\pm1\%$ & $22.1\pm0.7\%$ & $18\pm1\%$ & $21\pm1\%$ & $21\pm1\%$\\
&Index - LSTM & $25\pm1\%$ & $30\pm1\%$ & $30\pm1\%$ & $24\pm1\%$ & $23\pm1\%$
      & $21\pm2\%$ \\
&ViT - LSTM& $28\pm2\%$& $30\pm2\%$& $31\pm2\%$& $\mathbf{33\pm2\%}$ & $\mathbf{33\pm2\%}$ & $\mathbf{30\pm2}\%$ \\
&Aurora-T & $37\pm1\%$& $26\pm1\%$& $26\pm2\%$& $22\pm2\%$ & $22\pm1\%$ & $21\pm2\%$ \\
\bottomrule
\end{tabular}
}
    \end{center}
    \caption{Class-balanced accuracy over the test period (November $2012$ - March $2023$) for each lead week across persistence, climatology, hindcast, Logistic Regression, \networkA, \networkB, \networkC, and Aurora-T. For ML-based models, we report the mean and standard deviation across 100 trained models with varying random seeds.}
\label{tab:lead-week-acc}
\end{table}
\subsection{Forecast skill}\label{sec:fore-skill}

First, we evaluate the class-balanced accuracy for each lead week across our models and the baselines (Table \ref{tab:lead-week-acc}). For ML-based forecasts, we report the mean and standard deviation. The hindcast consistently outperforms all baselines in the first three weeks, while \networkC\ achieves comparable or superior performance for lead weeks $4$–$6$. Although medium-range forecasting is not our focus here, we note that the ML models trained solely on regime data (\networkA\ and LR) or those designed for shorter-term forecasting (Aurora-T) exhibit higher skill in the first two lead weeks but decline sharply beyond week $2$. In contrast, \networkB\ and \networkC\ exhibit increasing skill after lead week $1$, suggesting that the networks extract and learn long-range dynamical signals from these external drivers. However, the performance of \networkB\  decreases after lead week $3$ indicating that incorporating only SPV and MJO indices is insufficient for capturing teleconnection information on longer lead times.
The superior performance of our proposed \networkC\ in weeks 4-6 indicates that the ViT-based encoding of u10- and olr fields enhance the representation of atmospheric variability, improving teleconnections beyond the dynamics captured by SPV and MJO phase indices and thus increasing the robustness of long-term forecasts.

\begin{figure}[t!]
\includegraphics[width=\textwidth]{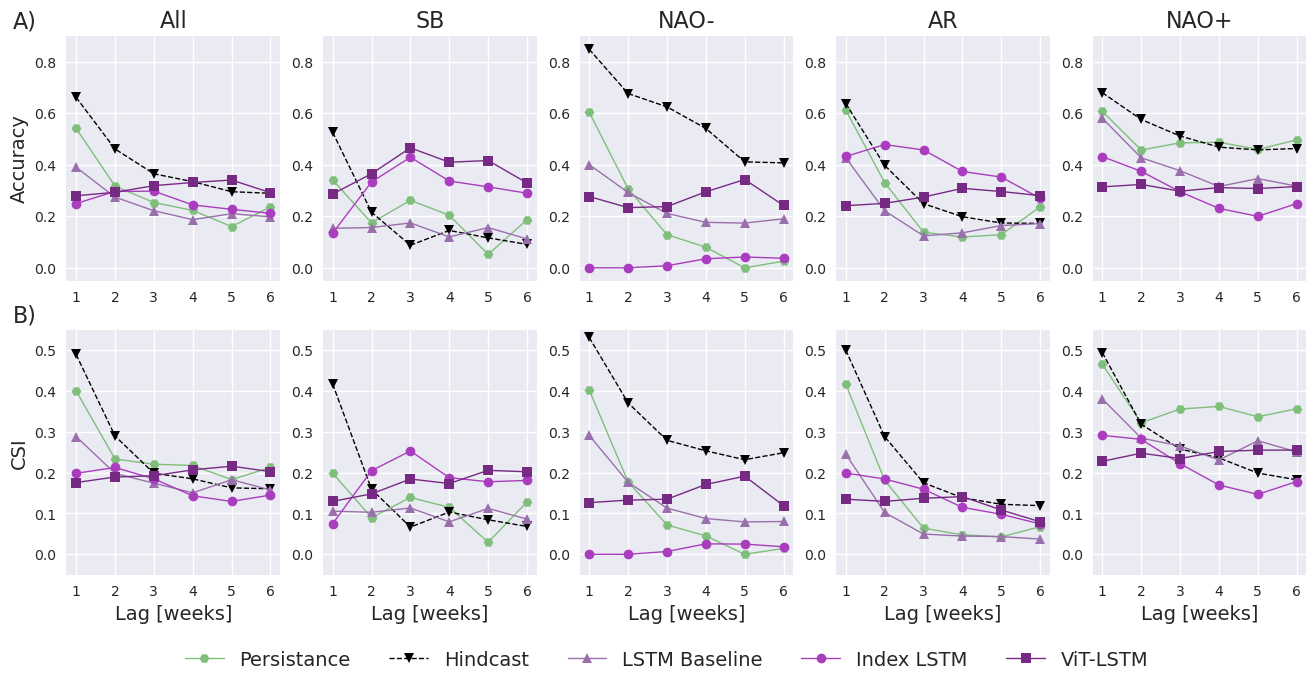}
    \caption{Analysis of class-wise mean forecast skill across regimes (different subplots with combined performance in the first plot) and predicted lead weeks (x-axis) using A) Accuracy and B) CSI (Critical Success Index)}
    \label{fig:acc-classwise}
\end{figure}

To better understand the differences in accuracy, we evaluate the class-wise performance of different models and baselines. The results for the balanced accuracy (top row) and Critical Success Index (CSI; bottom row) are shown in Figure \ref{fig:acc-classwise}. The first panel in each row presents the multi-class scores, while the remaining panels show regime-specific evaluations across the lead weeks. We focus on \networkA, \networkB, and \networkC\ alongside Persistence and the hindcast, excluding LR and Aurora-T, as they exhibit similar performance trends to LSTM and lack skill at the \stows-scale. Overall, the accuracy and CSI reveal mostly consistent class-wise forecast skill relationships, although larger discrepancies arise for the SB, AR, and the NAO$+$ performance of the hindcast. 
For the SB regime, both \networkB\ and \networkC\ outperform all models beyond lead week $1$, indicating a relevant role of teleconnection drivers in capturing long-term regime variability. While \networkC\ maintains higher accuracy values, likely due to its ViT-based encoding of u10 and olr fields, \networkB\ achieves a higher CSI value from lead weeks $2$ to $4$. 

For NAO$-$, the hindcasts remain the strongest performing method across all lead weeks. However, \networkC\ surpasses all other baselines beyond week 2, while \networkB\ exhibits the lowest forecast skill, except in lead weeks $5$ and $6$, where it outperforms persistence.

For AR, persistence and \networkA\ exhibit similar CSI and accuracy skill scores, indicating that \networkA\ lacks additional predictive signals beyond past regime information. \networkB\ demonstrates a significant performance gain in accuracy when classifying AR from lead week $2$ onward, whereas the CSI results show that the hindcast (all lead weeks) and \networkC\ (beyond lead week $3$) provide improved AR forecasts. The generally low CSI values for AR across all models indicate difficulties in correctly identifying true positives.

Regarding NAO$+$, the accuracy of the models indicates that the hindcasts and the persistence provide the most reliable classifications, while \networkB\ exhibits the lowest skill beyond lead week $3$ for both accuracy and CSI. \networkC\ exhibits the second lowest performance, however, for the CSI rankings, \networkC\ surpasses the hindcasts beyond lead week $3$, indicating improved identification of true positives at longer lead times. 
Despite maintaining high accuracy, the hindcasts struggle to detect NAO$+$ correctly from lead week 4 onward, as suggested by its declining CSI values.
Notably, the persistence scores remain stable across all lead weeks when classifying NAO$+$ for both accuracy and CSI, whereas the scores drop for all other regimes beyond lead week $2$. 
This strong NAO$+$ persistence could contribute to improved \stows\ predictions but is only subject to limited discussion in the literature \citep{Wu2022}, to the best of our knowledge (see Appendix \ref{app:skill} for details).

Overall, both CSI and accuracy indicate that the \networkA, which relies solely on past regime sequences and lacks information about remote drivers, rapidly loses predictive skill across nearly all regimes after lead week $1$, except NAO$+$, (e.g. ACC$(AR,t+1)\sim 42\%$), which is consistent with the persistence of this regime. While \networkB\ initially exhibits lower accuracy (ACC$(AR,t+1)\sim 35\%$), we find an improved \stows\  forecast skill beyond lead week $2$, except for the predictions of both NAO phases. The performance of \networkC\ indicates a behavior similar to that of \networkB, but consistently outperforms it, except for accuracy in AR predictions. This further underlines that while incorporating external driver information improves the model's ability to learn long-term dynamics, the flexible encoding of full climate fields, as in \networkC, provides additional improvements beyond the information captured by pure SPV and MJO phase indices.

\subsection{Windows of forecasting opportunity }\label{sec:woo}
To improve our understanding of the role of teleconnections and dominant climate patterns in \stows\ predictability, we analyze the association of skill with large-scale teleconnections and persistent NAE circulation patterns for our three LSTM architectures.
Specifically, we examine the states and temporal evolution of the NAE regimes, SPV, and MJO, preceding forecasts of enhanced predictability. We define such forecasts of opportunity as high-confidence predictions, that is when the model assigns a high probability to the predicted regime \citep{mayer_subseasonal_2020}. Because neural networks can be miscalibrated, meaning that the probabilities at the output do not align with the certainty of the prediction, we explicitly calibrate the probability outputs using a calibration loss term (see Section \ref{sec:training} and Appendix \ref{app:training}) to ensure that predicted probabilities accurately reflect model confidence \citep{mukhoti2020calibrating}. We then create a model ensemble by training $100$ models with varying random seeds and select only correct predictions within the 90th percentile and above, corresponding to at most $\geq16\pm1\%$ of the test samples (see also Appendix \ref{app:additional-results} and \citet{mayer_subseasonal_2020}).

\paragraph{Influence of preceding NAE regimes}

\begin{figure}[t!]
\includegraphics[width=\textwidth]{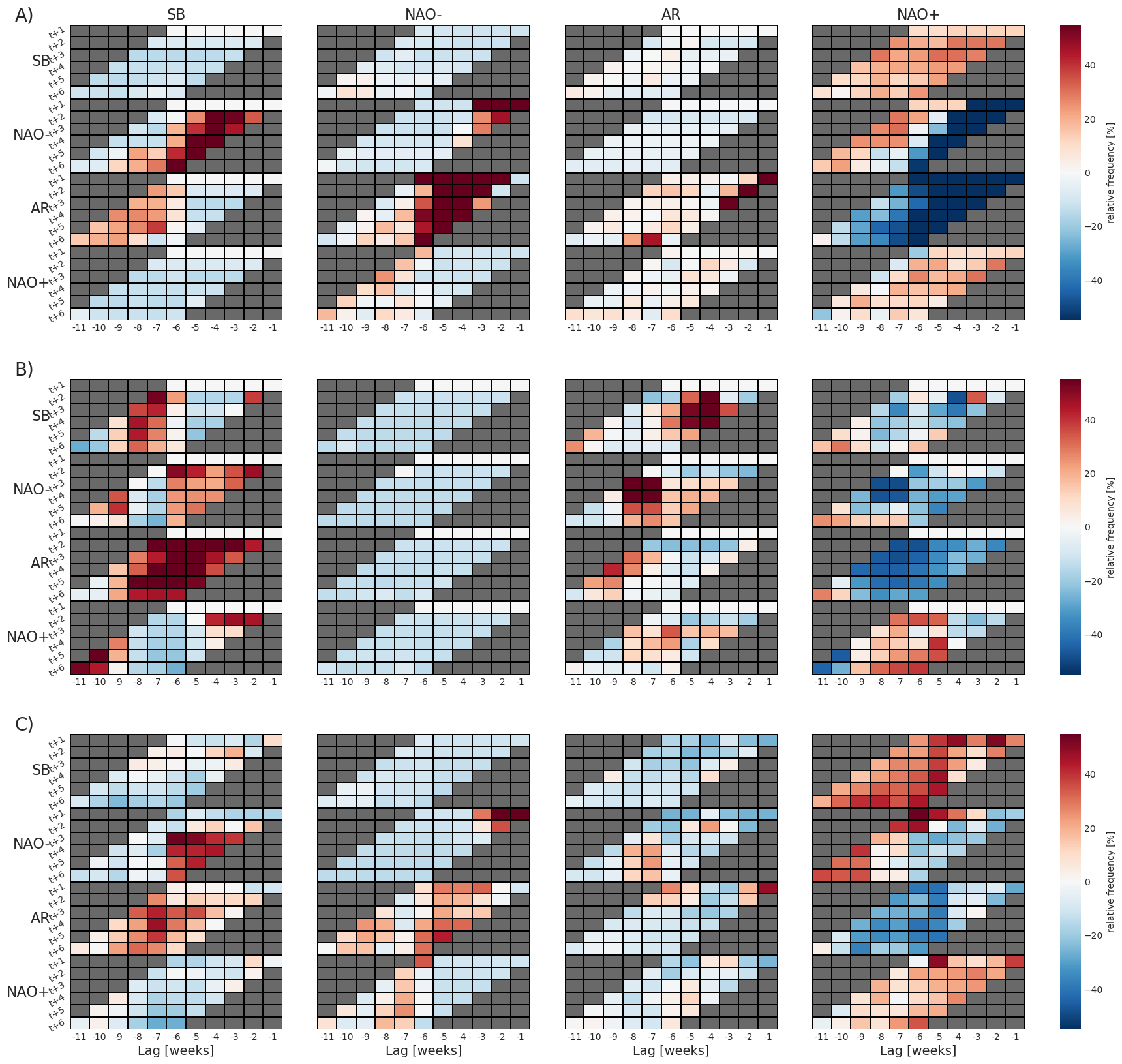}
    \caption[]{Analysis of how past NAE regimes influence model prediction probability of the regimes at different lead weeks.  In each panel, we can observe the relative prediction probability of regime B (subplot titles) at a specific lead week ($A,B \in \{\text{SB}, \text{NAO}-,\text{AR},\text{NAO}+\}$), given a regime A (y-axis) occurred several weeks before the regime B was predicted (Lag on x-axis). The values represent the prediction probability of regime B relative to the average regime frequency of regime B, highlighting precursor regime occurrence patterns beyond the expected frequency. The x-axis indicates the number of weeks before regime prediction, while rows within each subplot reflect a one-week lag shift in regime history based on their increased lead week prediction. To enhance interpretability, input, and output weeks were expanded to eleven time steps, allowing a clearer view of long-term dependencies. Panels A, B, and C correspond to \networkA, \networkB, and \networkC, respectively.}
    \label{fig:nae-frequency}
\end{figure}
To understand how past NAE regimes influence the prediction of future regimes, we compute the relative conditional probability of a regime class $Y$ being predicted in a specific lead week given that regime $X$ occurred in one of the previous weeks. For instance, we investigate questions such as: ``If regime SB occurred in input week $t-1$, how frequently is NAO$-$ predicted at lead week $t+2$ compared to its average occurrence?''. To understand the anomalies of the predicted regimes relative to the model's overall regime prediction frequency, we subtract the average occurrence probability of each regime across all samples. The results, grouped by model architecture \networkA\ - A, \networkB\ - B, and \networkC\ - C are shown in Figure \ref{fig:nae-frequency}. 
In each subplot, the y-axis represents the past regimes (SB, NAO$-$, AR, NAO$+$), while the x-axis corresponds to the number of weeks before prediction. The subplot titles indicate the predicted regime. Each row within a subplot represents a one-week shift in lag, illustrating how precursor regimes impact changes as lead time increases. For instance, in the SB subplot of \networkA\ (panel A), the first row shows the below-average probability of SB being predicted by the model in lead week 1 (t+1) one to six weeks after SB occurred. The second row shifts this perspective to lead week 2, reflecting the probability of SB being predicted two to seven weeks after SB onset, as one week more lies between the input weeks and the predictions of lead week 2. This pattern continues across all lead weeks, enabling an analysis of how past regime occurrences shape model predictions over time. Further details of the calculation methodology can be found in Appendix \ref{app:evaluation} and Equation \ref{eq:rel-freq}.

Across all three architectures, we observe consistent and inconsistent frequency patterns across the predicted regimes.

For SB forecasts, \networkC\ and \networkA\ exhibit similar precursor patterns, while deviating especially for medium-range lead times. In contrast, \networkB\ indicates increased SB frequency two to eight weeks after AR onset, while \networkC\ and \networkA show an increase in SB prediction probability two to six weeks after NAO$-$ occurred, with SB occurrence rising only six to eight weeks after AR onset. This difference in precursor patterns potentially contributes to \networkB's improved skill in lead weeks $2$ and $3$, while the lack of precursor patterns in lead week $1$ (row $t+1$) reflects the absence of high probability predictions.

For NAO$-$ forecasts, \networkA\ (A) and \networkC\ (C) capture a strong NAO$-$ occurrence ($>20\%$) one to three weeks after NAO$-$ occurred and an increased prediction probability ($>10\%$) two to six weeks after AR onset. In contrast, \networkB\ displays no clear pattern in the NAO$-$ column, which is consistent with its weaker forecast skill (Section \ref{sec:fore-skill}).

The AR predictions show few clear precursor regimes for \networkA\ and \networkC, with a small AR prediction probability increase ($>5\%$) one to three and seven weeks after AR onset. Meanwhile, \networkB\ associates increased AR predictions with an SB occurrence two to four weeks prior and NAO$-$ occurrence seven to eight weeks prior. A similar, less pronounced NAO$-$ pattern can be found in the \networkC\ results, which might contribute to the skill improvements for both networks.

For NAO$+$ predictions we observe the most similar patterns across the networks. In particular, \networkA\ and \networkC\ show strong alignment, reflecting their similar forecast skill. Nonetheless, all networks indicate a negative occurrence frequency (above-average absence) of NAO$+$, especially five to eight weeks following AR for \networkB\ and \networkC. However, while NAO$+$ is less frequent one to six weeks after NAO$-$ for \networkA\ and \networkC, \networkB\ results are less consistent. Similarly, though NAO$+$ occurs more frequently across all time lags for \networkA\ and \networkC, according to \networkB\ NAO$+$ is its own strongest precursor four to nine weeks after SB onset. Thus, aligning with \networkA\ and \networkC's worse performance compared to NAO$+$ persistence, in contrast to \networkB which performs closer to persistence skill.

\paragraph{Influence of the SPV}
\begin{figure}[t!]
\includegraphics[width=0.9\textwidth]{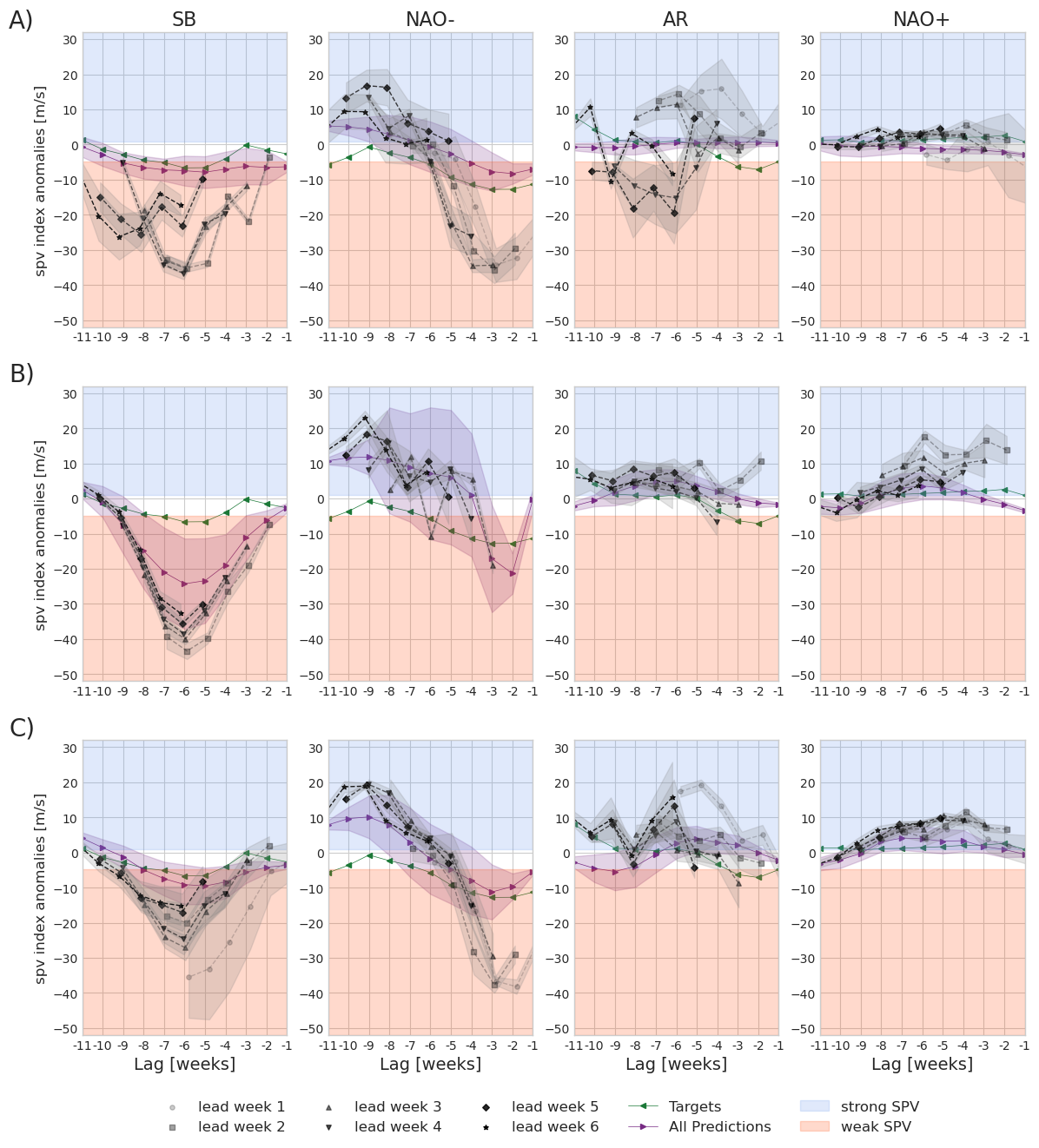}
    \caption[]{SPV index anomalies (relative to the mean spv index per class) over the lag in weeks for each predicted NAE regime (columns) for A) \networkA, B)\networkB, C)\networkC. The black dashed lines show the mean SPV index anomaly evolution across the six input weeks (different markers) for high-probability predicted regimes. The violet line represents the SPV anomalies across all regime predictions, that is true and false positives. The green line shows the SPV anomalies across all target events in the test set, coinciding for all models. Shading of the lines denotes standard deviation (across $100$ models). Strong vortex states, defined as exceeding the 80th percentile are indicated by the blue background shading, while weak SPV states, below the 30th percentile, are indicated in red \citep{Tripathi2015}.}
    \label{fig:SPV-index-anomalies}
\end{figure}
To further examine differences in forecast skill, we analyze and compare SPV index anomalies preceding network predictions of high probability from \networkA\ (A), \networkB\ (B), and \networkC\ (C) (Figure \ref{fig:SPV-index-anomalies}). Although \networkA\ lacks access to polar vortex information and \networkC\ includes spatial fields, we include all architectures in the analysis of the SPV index anomaly evolution to discern both large-scale teleconnections and the inherent associations between the NAE regimes and the SPV index. We show our results in Figure \ref{fig:SPV-index-anomalies}.

Both \networkB\ and \networkC\ show a significantly weakened polar vortex three to nine weeks before SB onset which then recovers to a neutral SPV state (black lines in the first panel of Figure \ref{fig:SPV-index-anomalies}). This U-shaped pattern is also visible as a weaker trend for all SB predictions (violet line) and is particularly pronounced for \networkB, which received the SPV index directly and achieved the second-highest skill for SB prediction after \networkC.  Interestingly, this is not visible in the SPV composites for all SB events (green line). This lack of a trend suggests that the SPV weakening, in the literature predominantly associated with NAO$-$, is also a long-lead predictor of SB, occurring after SPV recovery. Consistently, we find the NAO$-$ regime in \networkA and \networkC\ to precede SB predictions 2-6 weeks later (Figure \ref{fig:nae-frequency}). The similarly high SB prediction skill of \networkB\ and \networkC\ further indicates that the SPV index already contains the relevant stratospheric information for regime evolution. 

For NAO$-$, all networks indicate a strong SPV eight to eleven weeks before prediction, transitioning to a weak SPV one to four weeks prior (up to five weeks for \networkC). This pattern is not only visible for the high-probability predicted NAO$-$ regimes (black lines), but also for all NAO-prediction (violet lines), further being an amplification of the overall SPV evolution preceding NAO- regimes (green line). As noted above, a weak vortex preceding the NAO$-$ is also expected from the literature \citep{kretschmer2018more, domeisen_role_2020}. Nonetheless, we specifically observe a highly uncertain SPV index evolution preceding not only the high probability (black lines) but also all predictions (violet lines) of \networkB, which reflects the low NAO$-$ skill of this architecture. While this low-skill and less conclusive SPV pattern is somewhat surprising, it might be related to the strong association with the MJO (see next subsection). The similar skill, SPV evolutions, and regimes precursors (see the previous subsection) of \networkA and \networkC\  might further suggest that a weakening of the SPV together with a prevailing AR is followed by an NAO$-$. 

Similarly, the SPV evolution before AR events aligns with the NAE precursor pattern observed in Figure \ref{fig:SPV-index-anomalies}. While less continuous (see eight weeks prior) for \networkC, both \networkB\ and \networkC\ suggest a strengthened polar vortex eleven to five weeks before AR, consistent with AR’s tendency to occur three to six weeks before NAO$-$, which itself exhibits a strong SPV eleven to seven weeks earlier. \networkA\ provides no clear pattern, in line with its lower forecasting skill for AR.

For NAO$+$ and consistent with the literature, both \networkB\ and \networkC\ indicate a strengthened SPV three to eight weeks before onset, contradicting the earlier finding that SB serves as a precursor to NAO$+$, as SB is associated with a weak SPV during that time. Although other links might be present, further analysis is required to confirm our findings.

Overall, our results indicate a mixed picture regarding the role of the SPV in long-lead regime predictions. While including stratospheric information boosts skill beyond week 2 for SB and AR (Figure \ref{fig:acc-classwise}), it is less conclusive for the two NAO regimes. For NAO$+$ it seems that stratospheric information is already fully contained in the regime occurrences (Figure \ref{fig:nae-frequency}), consistent with the strong accuracy of a simple persistence forecast. For NAO$-$, \networkC\ outperforms \networkB\ suggesting that including the spatial stratospheric wind patterns provides information regarding a downward impact that the simple SPV index cannot.

\paragraph{Influence of the MJO}

\begin{figure}[t!]
\includegraphics[width=\textwidth]{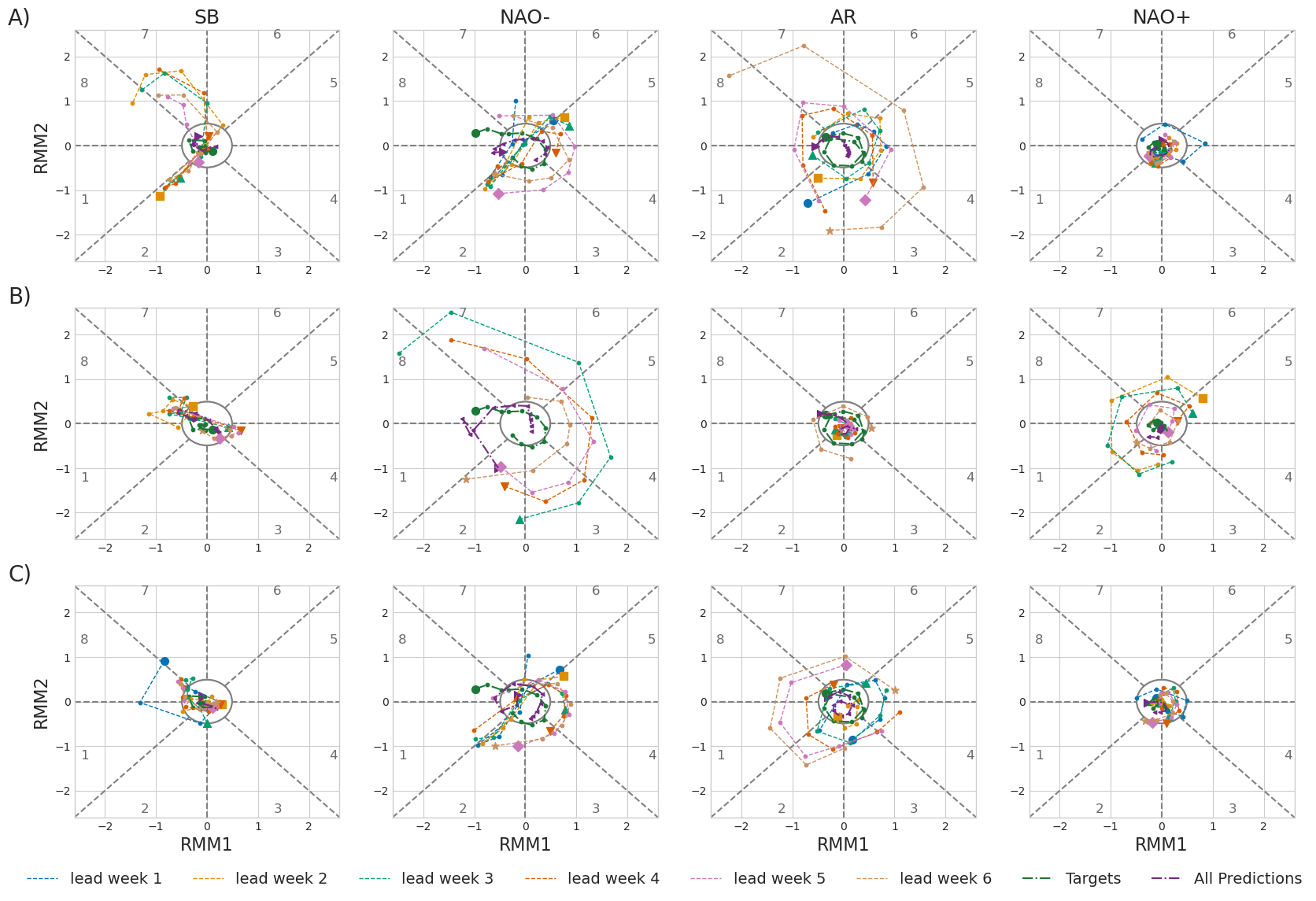}
    \caption[]{MJO phase diagram of two first principal components (RMM1 and RMM2) with phase evolution of the 6 input weeks according to the NAE regimes predicted for each lead week. The first input, i.e., $t-5$ is marked by the increased scatter point. A) \networkA, B)\networkB, C)\networkC}
    \label{fig:mjo-phase-diagram}
\end{figure}

Finally, we analyze MJO phase patterns to assess the influence of tropical variability on NAE regime forecasts. In Figure \ref{fig:mjo-phase-diagram}, we focus on the evolution of the MJO phase, represented by the behavior of the RMM indices in the phase space of the RMM 1 and RMM 2 components \citep{cassou_intraseasonal_2008,kiladis2014comparison}. Each line shows the evolution of MJO phase activity over the six input weeks. We highlight the first input week ($t-5$) with a large scatter point. The different lines indicate the six predicted lead weeks ($t+1$ to $t+6$). As a reference, we plot the average phase evolution of the model predictions (including correct and false) represented by the violet line, as well as the average phase evolution based on the target regimes (see Appendix \ref{app:additional-results}), represented by the green line. The unit circle (gray) in each plot indicates the amplitude threshold for active phases, set according to \citet{wheeler2004all, kiladis2014comparison} (taking the mean RMM and adding a standard deviation of the RMM, which is $~0.5$ for RMM1 and RMM2).

Across all regimes and architectures, we find that MJO phase activity before NAO regime predictions is more similar between \networkA\ and \networkC, despite their differing forecast skill. While AR precursor patterns indicate some similarities between \networkB\ and \networkC, consistent with their forecast skill, SB predictions exhibit highly variable phase evolutions across all architectures. The discrepancies between \networkB\ and \networkC\ suggest that the olr field in \networkC\ captures information beyond the MJO signal, thereby reducing the direct influence of the MJO phases on its predictions. 

For SB, both \networkA\ and \networkB\ indicate active MJO phase 8 around two weeks before prediction (yellow line), with both also showing activity in phases 8 and 7 at larger lags (three to eight weeks, green, orange, and pink lines). At time lags larger than ten weeks and four weeks, \networkB\ also suggests active phase 4, while \networkA\  indicates an active phase 7 between three and eight weeks before SB prediction. Contrary to the other architectures, \networkC\ indicates active phases 7, 1, and 2, occurring four to six weeks before SB onset (blue line). Additionally, we find that the average target and prediction phase activity lacks distinct phase precursors, indicating that the identified patterns may encode teleconnection signals.

For NAO$-$, all three architectures exhibit prominent cyclic phase patterns that persist across several lead weeks. This cyclic phase activity is consistent with the consecutive MJO phase shifts found at shorter timescales \citep{robertson_subseasonal_2020}. Specifically, all networks show sequential activations of phases 2, 3, 4, and 5 occurring eleven to seven weeks before NAO$-$ onset, with \networkC\ displaying this pattern for lead weeks 4 to 6 and \networkB\ for lead weeks 3 to 6. Additionally, active phases 6 and 7 emerge one to five weeks before NAO$-$ onset, as captured by either \networkC\ or \networkB\ (four and five weeks before), reinforcing learned features consistent with prior research \citep{Lee2019, lee2020links}.
Despite these overall similarities, the reliability of \networkB’s results is limited by its low forecast skill and low-probability predictions at shorter lead times (see Figure \ref{fig:90thperc} and Appendix \ref{app:additional-results}). Consequently, at shorter time lags, we primarily observe alignment between \networkA\ and \networkC, with both networks indicating an active phase 2 approximately two to five weeks before NAO$-$ onset. These findings, however, align with an increase in SB prediction probability two to six weeks after NAO$-$ onset, since we find MJO phase 2 activity four to six weeks preceding an SB prediction.

Similar to the NAO$-$, the AR precursors indicate consecutive MJO phase transitions before AR onset \citep{robertson_subseasonal_2020}. Especially, lead weeks 6, 5, and 4 (okra, pink, and orange) exhibit cyclic patterns for both \networkA\ and \networkC. Nonetheless, \networkC\ indicates a consistent sequence of phases 6, 8, 1, 2, and 3, which are active six to ten weeks before AR prediction, with phase 4 active five weeks prior. In contrast, \networkA\ shows a broader phase activation pattern (phases 2, 3, 4, 5, 7, and 8) six to eleven weeks before AR prediction, again almost consistently across lead weeks 6, 5, and 4 (okra, pink, and orange). Since, correct predictions from \networkA\ do not correspond to active MJO phases, unlike those from \networkB\ and \networkC\ (see Figure \ref{fig:activeMJOnumber} and Appendix \ref{app:additional-results}), MJO phases patterns of \networkC might be considered more influential. In addition, \networkC\ suggests phase 5 activity three to four weeks and eleven weeks before AR onset, aligning with earlier findings of AR occurring three to six weeks before NAO$-$, with NAO$-$ showing an active phase 5 seven to ten weeks before prediction. Nonetheless, the lack of resemblance between the patterns of \networkB\ and \networkC\ also leads to the assumption that the skill improvements for both \networkC\ and \networkB\  are not associated with MJO phase information.

NAO$+$ predictions exhibit inconsistent MJO phase evolution across all models, likely due to their low forecast skill. Prior research suggests strong MJO-NAE teleconnections at shorter lead times ($\leq15$ days) \citep{robertson_subseasonal_2020, Roberts2023}. Our results align with these findings, as the most active phases (for \networkA\ and \networkC) correspond to lags of up to four weeks. In addition, \networkB\ shows consecutive MJO phases between three and nine weeks prior and active phases $1$ to $3$ two to five weeks before onset in line with other findings \citep{Lee2019,lee2020links}. Nonetheless, further analysis is needed to confirm these patterns.

\section{Discussion}\label{sec:discuss}
Predictions beyond two weeks remain a fundamental challenge due to the chaotic nature of the climate system. Large-scale atmospheric teleconnections, such as the SPV and MJO, offer windows of enhanced predictability, which could significantly improve \stows\ forecasts. However, capturing these teleconnections with Machine Learning models remains a key challenge due to their multi-timescale dependencies and complex interactions with the climate system. Here, we address this challenge by developing deep learning architectures of increasing complexity, systematically evaluating how including teleconnection information influences \stows\ forecast skill of NAE regimes.

Our findings highlight the critical role of remote drivers in improving \stows\ predictability and a trade-off between short- and long-term predictability. Models that rely solely on regime sequences (\networkA, LR) lose forecast skill rapidly after short lead times due to their inability to capture the long-term influence of teleconnections. In contrast, architectures incorporating external climate fields---particularly \networkC, which integrates ViT-based encoding of u10 and olr fields, consistently improve long-range forecast skill, outperforming all models beyond lead week four. This suggests that incorporating encoded climate fields allows ML models to leverage teleconnections beyond existing climate indices, enhancing their ability to learn long-term dependencies. Interestingly, \networkC\ achieves forecast skill better than or comparable to the hindcast for almost all regimes, at the expense of limited NAO$+$ predictability, highlighting also potential improvement opportunities.

We further analyzed the dynamics that govern \stows\  forecast skill, by examining forecasts of opportunity, that is, high-confidence predictions that align with stable climate patterns. This analysis revealed distinct precursor relationships, shedding light on what the different networks learn. 
\begin{itemize}
    \item High-probability SB predictions are frequently preceded by AR (six to eight weeks prior) and NAO$-$ (two to six weeks prior). In addition, the increased SB forecast probability following NAO$-$ aligns with the strong-to-weak SPV transition preceding NAO$-$, coinciding with weak SPV phases before both regime forecasts, aligning with previous independent findings for both regimes \citep{kretschmer2017early, kretschmer2018more, spaeth2024stratospheric}. Similarly, SB and NAO$-$ display a common active MJO phase 2 as precursors (three to four weeks before SB and two to six weeks prior before NAO$-$). 
    \item NAO$-$ forecast probability increases one to three weeks after NAO$-$ occurred and two to six weeks after AR onset. The latter yields a promising teleconnection pattern, aligning with the coinciding strong SPV phases of NAO$-$ (eleven to seven weeks before prediction) and AR (eleven to five weeks prior). In addition, we find prominent cyclic MJO phase activity and active phases 6 and 7, one to five weeks before NAO$-$ onset, consistent with MJO patterns found at shorter timescales \citep{Lee2019, robertson_subseasonal_2020, lee2020links}.
    \item The precursor patterns of AR are limited. While AR forecast probability potentially increases eight to nine weeks after NAO$-$ and maintains persistence for up to three weeks, neither SPV patterns nor MJO phase activity confirm these findings. Nonetheless, similar to NAO$-$, we observe a stair-pattern sequence pattern in MJO phase activity preceding AR forecasts, aligning with prior findings \citep{Lee2019, robertson_subseasonal_2020}. 
    \item NAO$+$ forecasts are associated with an SPV strengthening before SB occurrence. However, the results remain inconsistent, mirroring the low skill across all ML models and leaving open questions regarding the NAO$+$ persistence as well as previously established MJO signals on longer time scales.
\end{itemize}

Overall, the results highlight that integrating teleconnections into \stows\ forecasts improves the forecasting skill, by providing long-term dynamical patterns. In addition, we demonstrated that \networkC\ benefits from encoded climate fields beyond conventional climate indices, enhancing long-range regime forecasts. While global numerical weather models still struggle to specifically account for such teleconnections, our findings indicate that Machine Learning approaches can offer more flexible and direct integration of external drivers. This also enabled us to further assess existing teleconnection indices and identify potential new teleconnection patterns, as demonstrated in our high-probability prediction analysis. These insights underscore the potential of physics-guided deep learning architectures to complement traditional forecast models, including \stows\ climate dynamics.

Despite our advances, several limitations and questions remain. For example, the assignment probabilistic NAE regimes have shown promise in capturing regime transitions more effectively \citep{spuler_identifying_2024}. Similarly, while u10 encoding improves the representation of tropospheric drivers, incorporating u-zonal wind at lower stratospheric levels could further enhance the learned stratosphere-troposphere interactions \citep{Baldwin2024}, potentially improving also NAO$+$ skill. In addition, accounting for inactive MJO phases in \networkB\ by passing the amplitude for each phase instead of categorical labels might improve MJO-related forecast performance. Finally, our analysis of learned patterns was based on input statistics, which limits interpretability and mechanistic understanding of the ML models.

To advance ML-based forecasting, future research could overcome these limitations by including probabilistic regime prediction to enhance regime forecast reliability and better capture regime transitions. Equally critical is the need for explainability and mechanistic interpretability \citep{mamalakis2020explainable,bommer2024finding}, enabling a true understanding of how ML models represent physical climate processes and ensuring that their skill improvements are rooted in atmospheric dynamics. 
By tackling these challenges, we can seamlessly integrate data-driven insights with physics-based forecasting, thereby transforming ML into a powerful tool for operational \stows\  predictions. This fusion of AI and climate science holds the potential to revolutionize \stows\ extreme weather forecasting, driving more reliable, interpretable, and actionable predictions.

\section*{Open Research Statement}
The source code for all experiments is accessible at (\href{https://github.com/philine-bommer/DL4S2S}{https://github.com/philine-bommer/DL4S2S}) and will be fully executable upon publication. All experiments and code are based on Python v3.10.6. All dataset references are provided throughout the study. 

\section*{Conflict of Interest declaration} 
The authors declare that they have no
affiliations with or involvement in any organization or entity with any
financial interest in the subject matter or materials discussed in this
manuscript.
\section*{Author Contributions} PB contributed to all parts of this work and the development of the code basis. MK and MH to development of the experiments and curation of the results. FS and KB contributed to the experiments and evaluation. All authors contributed to the writing of the manuscript.

\ack
This work was funded by the German Ministry for Education and Research through project Explaining 4.0 (ref. 01IS200551) and REFRAME (ref. 01IS24073B). M.K. acknowledges funding from XAIDA (European Union’s Horizon 2020 research and innovation program under grant agreement No 101003469).
NOAA/CIRES/DOE 20th Century Reanalysis (V3) data was provided by the NOAA PSL, Boulder, Colorado, USA, from their website at https://psl.noaa.gov. Support for the Twentieth Century Reanalysis Project version 3 dataset is provided by the U.S. Department of Energy, Office of Science Biological and Environmental Research (BER), the National Oceanic and Atmospheric Administration Climate Program Office, and the NOAA Earth System Research Laboratory Physical Sciences Laboratory. Subseasonal hindcasts were accessed through MARS, the ECMWF meteorological archive.
We also acknowledge the contribution of Paul Boehnke, who provided preliminary efforts and results in his Master's Thesis.

\newpage
\appendix
\section{Data}\label{app:data}
\subsection{Technical preprocessing}\label{app:methods-preprocessing-technical}
In Table \ref{tab:variables}, we provide a summary of all used climate variables. Since we used the first version of WeatherBench data\citep{rasp_weatherbench_2020} for preliminary architecture testing, we adapted a similar regridding resolution and scaled all variables to a 1.40525$^\circ$ grid.

As discussed in the main body, prior to training we standardize all input variables to a mean $\mu=0$ and standard deviation $\sigma=1$. In datasets with one-dimensional features, this normalization is done for each feature across all samples of the train dataset. In the case of the present work with time series of 2D-maps among the input features the normalization is applied across both spatial dimensions and the time dimension. This normalization is applied for each climate variable separately, at a grid point $(x, y)$  at time $t$: 
\begin{equation}
    Norm(X_{t,x,y}) = \frac{X_{t,x,y} - \mu(X)_{x,y}}{std(X)_{x,y}},
    \label{eq:std-norm}
\end{equation}
with 
\begin{equation}
    \mu(X)_{x,y}= \frac{\sum_t^T{X_{t,x,y}}} {T},
\end{equation}
and
\begin{equation}
    std({X})_{x,y} = \sqrt{\frac {sum_t^T{(X_{t,x,y} - \mu(X)_{x,y})^2} }  {T} }.
\end{equation}
The length of the time series is denoted by $T$. \\
While both the weather regime time series and the MJO phase index, as binary class vectors do not require normalization, we normalize the real-valued SPV index by subtracting the mean and dividing by the standard deviation, similar to Equation \ref{eq:std-norm}.

\begin{table}[h!]
\centering
 \resizebox{\textwidth}{!}{ 
\begin{tabular}{llll}
\toprule
        \textbf{Variable name} & \textbf{Region} & \textbf{Unit} & \textbf{Levels} \\
        \midrule
        Geopotential Height (z) & $90^{\circ}$W–$30^{\circ}$E, $20^{\circ}$–$80^{\circ}$N & $m$ & $500\text{ hPa}$\\
        \midrule
        SPV index \citep{domeisen_role_2020}& $60^\circ$N &  $\text{ms}^{-1}$ &  $10\text{ hPa}$\\
        MJO phase index \citep{wheeler2004all} & $15^\circ\text{S} - 15^\circ$N & multiple & multiple\\
        \midrule
        U component of wind (u) & $60^\circ - 90^\circ$N &  $\text{ms}^{-1}$ & $10\text{ hPa}$\\
        Outgoing longwave radiation (olr) & $15^\circ\text{S} - 15^\circ$N & $Wm^{-2}$ & - \\
        \bottomrule
    \end{tabular}}
    \label{tab:variables}
    \caption{Weather variables used in this work.}
\end{table}

\section{Methods}
To ensure the reproducibility of all results and architectures, in the following, we provide additional details on all used architectures, training, hyperparameters, and evaluation steps.
\begin{figure}
\includegraphics[width=\textwidth]{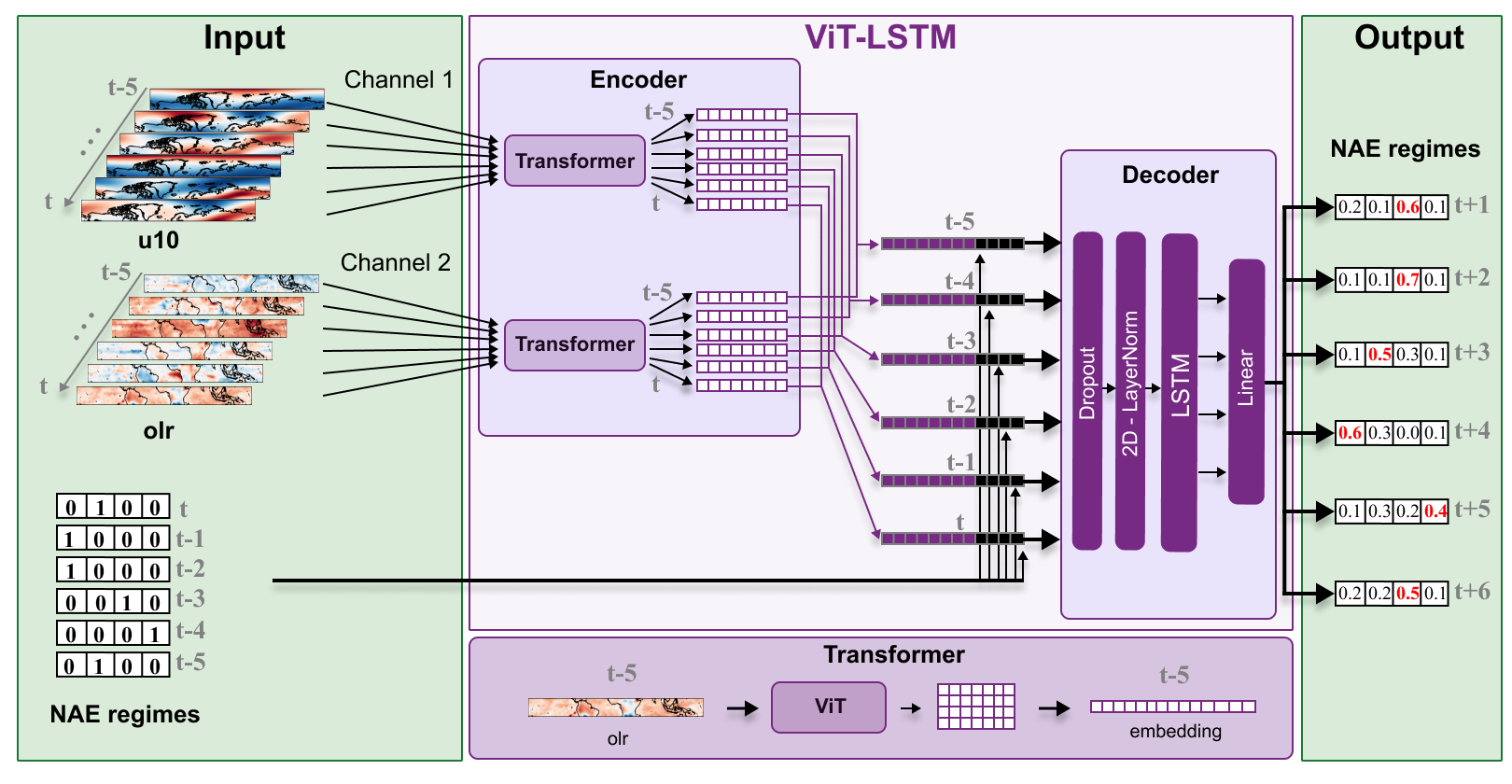}
    \caption[]{Schematic of the ViT-LSTM architecture, with detailed encoder and decoder layout.}
    \label{fig:A-vit-lstm}
\end{figure}

\subsection{LSTM \& Index-LSTM}\label{fig:A-preprocessing}
As depicted in Figure \ref{fig:A-vit-lstm} in the decoder panel, our \networkA\ (same as \networkB\ and the decoder of \networkC) consists of a single LSTM layer with the number of hidden states (see Table \ref{tab:hps}) determined during hyperparameter optimization. The LSTM layer is followed by a time-distributed linear layer which predicts the probability of each class for the next $6$ weeks.

\subsection{ViT-LSTM architecture}\label{app:architectures}
The encoder consists of two ViTs that process individual time steps of either the u10 (Channel 1 in Figure \ref{fig:A-vit-lstm}) and olr fields (Channel 2 in Figure \ref{fig:A-vit-lstm}). Thus, each climate variable field (image) of the previous 6 weeks is passed to the corresponding ViT, mapping each time step of olr and u10 to an embedding space of size $d_{\epsilon} = 32$ (output before last linear layer). To handle the limited training data (see Sec.~\ref{sec:data}), we apply dropout \citep{srivastava2014dropout} and a batch normalization layer \citep{ioffe2015batch} to the embeddings.

Next, the generated embeddings (violet array in Figure \ref{fig:A-vit-lstm}) are normalized by subtracting the minimum and dividing by the difference between maximum and minimum. Then, we combined the embeddings with the binary 4-dimensional NAE regime class information of the previous 6 weeks (black-outlined array). Each time step results in a one-dimensional vector, which is standardized. These new vectors contain information from both the NAE regimes and the olr and u10 data. The combined vectors are passed first through a Batch Normalization and a dropout layer, before being passed through the decoder. The \networkA\ decoder follows the same architecture described in Section \ref{sec:lstm-network} and the output is the 2-dimensional time series shown in the output panel in Figure \ref{fig:A-vit-lstm}.

\paragraph{Transformer \& MAE training}
To pre-train our ViT encoder, we first build an MAE for each climate variable following \citet{He2022}. An MAE is a self-supervised deep learning model that reconstructs missing (masked) parts of an image (either the u10 or olr field), by learning from the surrounding image parts. During training, the image is segmented into patches (see patch size in Table \ref{tab:mae-hp}) and a large portion of the patches is randomly masked. The model is trained to predict the missing content using an encoder-decoder architecture, where the encoder (here a ViT), encodes only visible data and the MLP decoder reconstructs the missing parts based on the encoder embeddings. This forces the model to learn meaningful high-level features, making it useful for tasks like representation learning and transfer learning.

In our \networkC\ we then use the MAE encoder part, by reconstructing only the encoder ViT (see Table \ref{tab:mae-hp} for several attentions heads and other architecture details) with the parameters of the best trained MAE model. The ViT architecture follows \citet{dosovitskiy2020image} and was implemented based on the PyTorch \textit{vit-pytorch} package. 

\subsection{Hyperparameters}\label{app:training}

All architectures are trained using the ADAM optimizer \citep{kingma2014adam}. While smaller architectures (\networkA\ and \networkB) are less prone to miscalibration, \networkC\ does not return calibrated probabilities. Nonetheless, to ensure proper calibration, we train all architectures using an adaptive Focal Loss function \citep{mukhoti2020calibrating}.
To reduce overfitting, we apply the early stopping technique, which works by stopping training early once a predefined metric, i.e., accuracy, stops improving on the validation set. We also applied gradient clipping \citep{pascanu2013difficulty}, Stochastic Weight Averaging (SWA)\citep{izmailov2018averaging}, and an L2-regularization (weight decay). As discussed in the main body material, we use Bayesian Optimization (BO) to select the optimal hyperparameter set. BO is a probabilistic method that efficiently optimizes unknown but costly functions by using a surrogate model, like a Gaussian Process, to guide sampling through an acquisition function. Used in hyperparameter optimization, BO tends to be more efficient than direct searches, such as grid search or random search. Due to the network similarity (see Section \ref{sec:methods}) and to limit computational cost, we perform the BO on the classification setting of \networkC\ (see Section \ref{sec:training}) and adopt the same hyperparameters for all LSTM-based architectures. We arrive at the following setup.  

\begin{table}[h!]
\begin{center}
    \resizebox{\textwidth}{!}{ 
\begin{tabular}{lllll}
\toprule
\textbf{Hyperparameter} &\textbf{\networkA} & \textbf{\networkB} & \textbf{\networkC} \\
\midrule
hidden states (LSTM/Decoder) & $256$& $256$ & $256$ \\
dropout& $0.165$ & $0.165$ & $0.165$  \\
learning rate & $0.0001$ & $0.0001$ & $0.0001$ \\
batch size & $72$ & $72$ & $72$\\
weight decay & $0.0009$& $0.0009$ & $0.0009$ \\
Gradient clipping& $0.827$ & $0.827$ &$0.827$ \\
SWA & $2.5\times10^{-5}$ & $2.5\times10^{-5}$ & $2.5\times10^{-5}$ \\
\midrule
\end{tabular}
}
    \end{center}
    \caption{Hyperparameters determined via BO with $n_b = 100$ steps, maximizing the average validation accuracy $A_{\mathrm{val}} = 32.7\%$)}
    \vskip -0.1in
\label{tab:hps}
\end{table}

\paragraph{MAE}
As described in Section \ref{sec:training}, we train on a combined training dataset consisting of the 20CRv3 data between $1836$ and $1969$ and ERA5 between $1980$ and $2009$. As validation data, we use the last 10 years of 20CRv3 data (November 1970 until March 1980). Due to the computational cost of the MAE training, we chose the best hyperparameters based on the lowest validation reconstruction error across three model configurations. The corresponding hyperparameters of the MAE and ViT-encoder are provided in Table \ref{tab:mae-hp}.
\begin{table}[h!]
\begin{center}
\begin{tabular}{lllll}
\toprule
\textbf{Hyperparameter} &\textbf{ViT} & \textbf{MAE} & \\
\midrule
channels & $1$&$-$\\
depth & $6$&$-$\\
dim & $512$&$-$\\
dropout & $0.1$&$-$\\
embedding dropout & $0.1$&$-$\\
attention heads & $16$&$-$\\
MLP dimension& $2048$&$-$\\
patch size & $2\times16$&$-$\\
\midrule
decoder depth & $-$& $6$\\
decoder dimension &$-$ &$32$\\
masking ratio &$-$ &$0.75$\\
\midrule
\end{tabular}
    \end{center}
    \caption{Hyperparameters determined across three configurations with the smallest validation reconstruction error ($E_{rc}\leq0.1$)}
    \vskip -0.1in
\label{tab:mae-hp}
\end{table}

\subsection{Additional Models}\label{app:add-models}
\paragraph{Logistic regression}
Due to the definition of the LR (based on sklearn implementation), we have to predict each week individually. However, to maintain comparability, the same LR model should predict all six lead weeks. While we tested training individual models for each lead week with no change in performance, we designed the LR model to predict each lead week individually, but always based on the same six input weeks. In other words the input for lead week $t+6$ is the same as for the prediction of lead week $t+1$, i.e., the past six weeks ($t-5$ to $t$).

\paragraph{Aurora - T}
Given the success of climate foundation models for medium-range predictions, we tried to adapt such an architecture to the \stows\ timescale.
We chose Aurora due to its easy access and well-documented code \citep{bodnar2024aurora} and used the large pre-trained version for higher-resolution data. 
As input the model receives the extended northern hemisphere ($15^\circ$S$-90^\circ$N) of u10 as the atmospheric variable and the extended northern hemisphere of olr as a surface variable. While Aurora is only pre-trained on atmospheric data up to $50$hPa, we argue that the u-zonal wind dynamics of $50$ and $10$hPa show close resemblance. Thus, we pass the u10 field in place of the u-zonal wind at $50$hPa (Since we have to provide the height information as input for Aurora). Nonetheless, we point out that these foundation models are known to struggle with stratospheric data. 
As Aurora embedding we use the activations of the backbone model \citep{bodnar2024aurora}. Thus, each embedding vector includes information for two weeks (Aurora is trained with two input timesteps) for both u10 and olr fields. To reduce the complexity and computational cost, we apply PCA and maintain the first three PCs (capturing $95\%$ of the explained variance). The PCs are then collected for the six input timesteps to create the image embedding vector. The image embedding vector and the nae regime classes of the last six weeks are then passed to a temporal transformer, consisting of two transformer encoder layers (self-attention layer) --for the image embeddings and one for the regimes-- followed by a transformer decoder layer. The transformer decoder layer connects the output of the two prior transformer layers to predict the probabilities of the next six weeks.  

\subsection{Evaluation}\label{app:evaluation}
For the class-wise accuracy, we calculate the accuracy as defined for a binary classification scenario:
\begin{align}
    \mathrm{Accuracy} = \frac{\mathrm{TP} + \mathrm{TN}}{\mathrm{TN}+\mathrm{TP}+\mathrm{FP}+\mathrm{FN}}
    \label{eq:A-acc}
\end{align}
where $\text{TN}$ is the number of true negatives, $\text{TP}$ the number of true positives, $\text{FP}$ the number of false positives and $\text{FN}$ the number of false negatives.

\paragraph{NAE precursors}
To calculate the relative frequency we first calculate the conditional probability $p(x|y)$ that an NAE regime $k \in C$ was predicted in the output $y_{n,[t+i]},i\in[1,6]$, given an NAE regime $c\in C$ occurred in input $x_{n,[t-j]}, j \in[0,5]$ across all high probability predictions $N$(above 90th percentile), with $i,j\in\mathbb{N}$. As our reference probability, we also calculate the probability $p(y)$ that an NAE regime $k \in C$ occurred in the ground truth output across all samples $M$ in the test set. All probabilities can be defined, using an indicator function $\mathbb{I}:\mathbb{R}\mapsto\{0,1\}$, as:
\begin{align}
    p(x\cap y)_{k,c,i,i-j+1} &= \frac{1}{N}\sum_{n=1}^N\mathbb{I}(y_{n,[t+i]}=k)\mathbb{I}(x_{n,[t-j]}=c),\\
    p(x)_{c,i,i-j+1} &= \frac{1}{C}\sum_{k=0}^C p(x\cap y)_{k,c,i,i-j+1},\\
    p(y)_{k,i,i-j+1} &= \frac{1}{C}\sum_{c=0}^C \frac{1}{M}\sum_{m=1}^M\mathbb{I}(\hat{y}_{m,[t+i]}=k)\mathbb{I}(x_{m,[t-j]}=c)\\
    \label{eq:app-occ}
\end{align}
with $x_n\in\mathbb{R}^{1\times 6}$ being the regime input vector of the n-th input sample, $y_{n}$ being the predicted class labels across all lead weeks for the sample, and $\hat{y}_{m}$ the m-th target regimes according to ERA5 of all lead weeks. Correspondingly, $x_{n,[t-j]}$ is the regime in input week $t+i$ of a sample $n$, $y_{n,[t+i]}$ is the predicted regime label and $\hat{y}_{m,[t+i]}$ in lead week $t+i$. 
Based on Equation \ref{eq:app-occ}, we we then define the conditional probability $p(y=k|x=c)$ of regime $k$ being predicted in lead week $t+i$, given regime $c$ occured $dt = j-i+1$ weeks before, as follows:
\begin{align}
    p(y=k|x=c)_{k,c,i,i-j+1} = \frac{p(x\cap y)_{k,c,i,i-j+1}}{p(x)_{c,i,i-j+1}}.
    \label{eq:app-condP}
\end{align}
Thus, we derive the relative frequency $\bar{\mathbf{f}} \in \mathbb{R}^{C\times C\times T \times 2T-1}$ from \citet{cassou_intraseasonal_2008} as:
\begin{align}
    \bar{f}_{c,k,i,i-j+1} &= p(y=k|x=c)_{k,c,i,i-j+1} - p(y)_{k,i,i-j+1},
    \label{eq:rel-freq}
\end{align}
indicating a value for occurrence anomalies relative to the climatological regime occurrence.

\paragraph{Hindcast}
To calculate the performance of the hindcast, we define the lead weeks predictions of the hindcast at day $6$ (lead week $1$), $13$ (lead week $2$), $20$ (lead week $3$), $27$ (lead week $4$), $34$ (lead week $5$) and $40$ (lead week $6$). This layout was chosen due to the format and extent of the data. Furthermore, we calculate all forecast skill metrics across the full range of the forecast (i.e., $1980-2020$). We argue that this data range provides comparable statistics to our ERA5 test data range since the hindcast is initialized only once a week leading to $~24$ initializations per year in boreal winter. 

\section{Additional Experiments}\label{app:additional-results}
In this section, we provide additional results and discuss additional findings. The results of all results presented here follow the procedure outlined in the main body material.

\subsection{Forecast skill}\label{app:skill}
While the most relevant results are discussed in the main body, in the following we focus on further insights gained from the forecast skill analysis and the corresponding implications.
\paragraph{Persistence}
In both Table \ref{tab:lead-week-acc} and Figure \ref{fig:acc-classwise}, the persistence skill stands out in that it outperforms all models and baselines except for the hindcast in lead week $1$ and in lead week $2$ together with the LR model. While these results are in line with the established strong persistence of NAE regimes on shorter timescales (e.g. see \citet{nielsen_forecasting_2022}), the strong persistence of the NAO$+$ even on longer time scales has only been a novel but limited research focus, to the best of our knowledge \citep{Wu2022}. Both accuracy and CSI scores suggest superior performance of the persistence forecast after lead week $3$, with the CSI suggesting the overall highest true positive rate after lead week $2$. Though the accuracy results suggest that the hindcast NAO$+$ skill might be based on predicting a persistent NAO$+$ after lead week $3$, the CSI score does not support this assumption. Thus, the relation between hindcast and NAO$+$ persistence, as well as the overall NAO$+$ persistence requires further investigation. Nonetheless, we point out that the persistence of NAO$+$ between lead week $3$ to $6$, could give rise to an improved \stows\ forecast, as NAO$+$ causes for example higher winter precipitation accompanied by higher temperatures over northern Europe and lower precipitation with higher temperatures over the Mediterranean \citep{scaife2005stratospheric,Rousi2020}.

\paragraph{Logistic Regression \& Aurora-T}
Similar to \networkA, both the LR model and the Aurora-T model show high initial forecast skill in lead week $1$(compared to \networkB\ and \networkC), with drastically decreasing skill starting at lead week $2$ (except for LR in lead week $2$). For LR these results align with its architectural similarity to \networkA, since the input is limited to the NAE regime features of the past six weeks. However, for Aurora-T, the decrease in performance indicates that the embeddings of u10 and olr generated by the Aurora backbone (see Appendix \ref{app:add-models}) might limited to short-term dynamics due to the training on a medium-range weather forecasting task. In addition, models such as Aurora struggle with stratospheric data, further hampering the extraction of impactful external driver information from the u10 data. Thus, though outside of the scope of this work, we point out that improvements such as fine-tuning on stratospheric data or an \stows\  prediction range could drastically improve the forecast skill of Aurora-T.
\begin{figure}[t!]
\includegraphics[width=\textwidth]{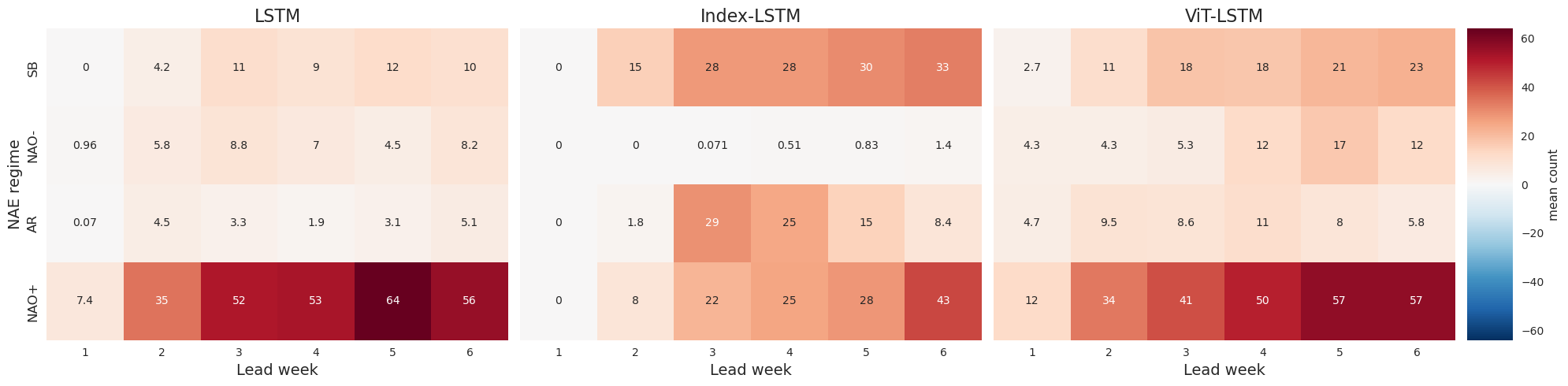}
    \caption[]{Occurences of correct predictions with probabilities in the 90th percentile per predicted regime and lead week.}
    \label{fig:90thperc}
\end{figure}
\subsection{Prediction patterns and external drivers}\label{app:external-driver-calc}
To support the analysis of forecasts of opportunity (as defined in the main body) and corresponding external driver impact, here we provide further statistical analysis.

\paragraph{Forecasts of opportunity}
As discussed in Section \ref{sec:results}, we define a forecast of opportunity as a prediction with a certainty above the 90th percentile. Due to accuracy and CSI variations across predicted lead weeks and regimes (see Table \ref{tab:lead-week-acc} and Figure \ref{fig:acc-classwise}) the number of forecasts of opportunity varies not only across models (\networkA, \networkB, and \networkC). Thus, in Figure \ref{fig:90thperc} we plot the mean occurrence of 90th percentile predictions per regime and timestep across the deep ensemble (i.e., $100$ trained network). Each cell is annotated by the mean count. The results align with the forecast skill results (see Table \ref{tab:lead-week-acc} and Figure \ref{fig:acc-classwise}) and further demonstrate that our networks are well-calibrated.

\begin{figure}[t!]
\includegraphics[width=\textwidth]{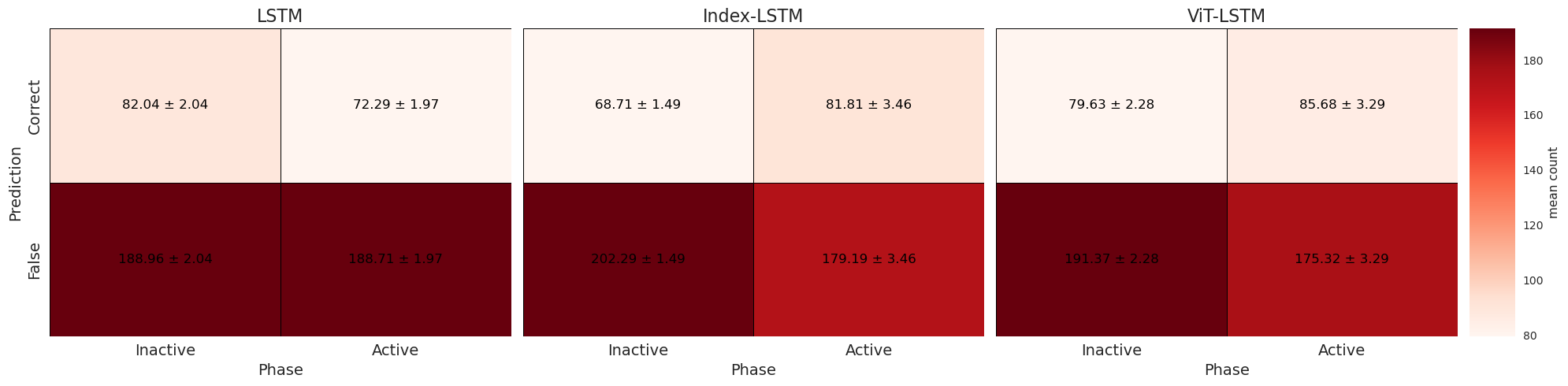}
    \caption[]{Occurences of correct or incorrect predictions with prior active or inactive MJO phase. The annotations in each box show the mean and standard deviation.}
    \label{fig:activeMJOnumber}
\end{figure}

\paragraph{MJO phases}
For the analysis of the MJO phases as precursors of forecasts of opportunity, we consider the average MJO phase activity across all network predictions see the violet line in Figure \ref{fig:mjo-phase-diagram} and the average phase activity across all targets, plotted as the green line. To calculate these two references, we collect the average RMM1 and RMM2 for each week before a predicted lead week, Thus, we extract the average RMM per predicted NAE regime $c$ for each time lag $dt =[1,..,11]$ weeks, as follows:
\begin{align}
 RMM^{\text{pred}}_{dt = i-j+1}(c) &=\frac{1}{N}\sum_{n=1}^{N}\mathbb{I}(y_{n,t+i} = c)RMM_{n,t-j},\\
 RMM^{\text{target}}_{dt=i-j+1}(c) &= \frac{1}{N}\sum_{n=1}^{N}\mathbb{I}(\hat{y}_{n,t+i} = c)RMM_{n,t-j},
 \label{eq:rmm-average}
\end{align}
with $N$ the number of samples, $y_{n,t+i}$ the predicted regime label in lead week $t+i$, and $\hat{y}_{n,t+i}$ the target regime label in lead week $t+i$. In Equation \ref{eq:rmm-average}, RMM can represent either RMM1 or RMM2 and $RMM_{n,t-j}$ is the RMM of the n-th input sample in input week $t-j$.
In addition, we consider only active MJO phases, as inactive phases have less impact on atmospheric conditions \citep{robertson_subseasonal_2020}. Thus, the relationship between active phases and correct predictions might provide further insight into learned teleconnections. Although only \networkB\ has direct information about the phase activity (inactive phases correspond to input class $0$, see Section \ref{sec:data}), we evaluate the relationship of the prediction correctness and phase activity across all three architectures. Figure \ref{fig:activeMJOnumber}, shows the confusion matrix of each model, with the rows indicating correct or false predictions and the columns indicating inactive or active predictions. Each cell is annotated by the mean count and the standard deviation across the deep ensemble.  
In line with the lack of external driver access, we find that \networkA\ does not indicate a higher number of correct predictions for active phases. In contrast, both \networkB\ and \networkC\ indicate significantly higher values of correct predictions for active phases, further demonstrating learned teleconnections. Nonetheless, we point out that \networkB\ does indicate a stronger relationship between correct predictions and active MJO phases. In combination with \networkB's overall lower performance, these results indicate, however, that the MJO phases alone are not the most skillful precursor in the tropics.

%

%
\newpage
  
\printbibliography

\end{document}